\def\year{2018}\relax
\documentclass[letterpaper]{article} 
\usepackage{aaai18}  
\usepackage{times}  
\usepackage{helvet}  
\usepackage{courier}  
\usepackage{url}  
\usepackage{graphicx}  
\usepackage{xcolor}

\graphicspath{{./}}
\usepackage{amssymb}
\usepackage{amsthm}
\usepackage{mathtools}
\usepackage{tabularx}
\usepackage{multirow}

\newcommand{\specialcell}[2][c]{%
	\begin{tabular}[#1]{@{}c@{}}#2\end{tabular}}
\newcommand{\etal}{\textit{et al}. }
\newcommand{\ie}{\textit{i.e.} }
\begin{document}
	\title{Visual Question Generation as Dual Task of Visual Question Answering}
	\author{Yikang Li$^{1}$, Nan Duan$^{2}$, Bolei Zhou$^{3}$, Xiao Chu$^{1}$, Wanli Ouyang$^{4}$, Xiaogang Wang$^{1}$\\ 
		$^{1}$The Chinese University of Hong Kong, Hong Kong, China \ \ \ \ \ $^{2}$Microsoft Research Asia, China\\ 
		$^{3}$Massachusetts Institute of Technology, USA  \ \ \ \ \  $^{4}$University of Sydney,  Australia\\
	}

	\maketitle
	\begin{abstract}
		Recently visual question answering~(VQA) and visual question generation~(VQG) are two trending topics in the computer vision, which have been explored separately. In this work, we propose an end-to-end unified framework, the Invertible Question Answering Network~(iQAN), to leverage the complementary relations between questions and answers in images by jointly training the model on VQA and VQG tasks. Corresponding parameter sharing scheme and regular terms are proposed as constraints to explicitly leverage $\langle$Q,A$\rangle$'s dependencies to guide the training process. After training, iQAN can take either question or answer as input, then output the counterpart. Evaluated on the large scale visual question answering datasets CLEVR and VQA2, our iQAN improves the VQA accuracy over the baselines. We also show the dual learning framework of iQAN can be generalized to other VQA architectures and consistently improve the results over both the VQA and VQG tasks.\footnote{Source code will be released when accepted.}
		
	\end{abstract}
	
	\section{Introduction}
	
	
	Question answering~(QA) and question generation~(QG) are two fundamental tasks in natural language processing~\cite{manning1999foundations,martin2000speech}. In recent years, computer vision has been introduced so that they become cross-modality tasks, Visual Question Answering~(VQA)~\cite{zhou2015simple,yumulti,antol2015vqa} 
	and Visual Question Generation~(VQG)~\cite{mostafazadeh2016generating,zhang2016automatic}, which refer to the techniques of the computer vision and the natural language processing. 
	Both VQA and VQG tasks involve reasoning between a question text and an answer text based on the content of the given image. The task of VQA is to answer image-based questions, while the VQG aims at generating reasonable questions based on the image content and the given answer. 
	
	\begin{figure}[t]
		\begin{center}
			\includegraphics[width=0.9\linewidth]{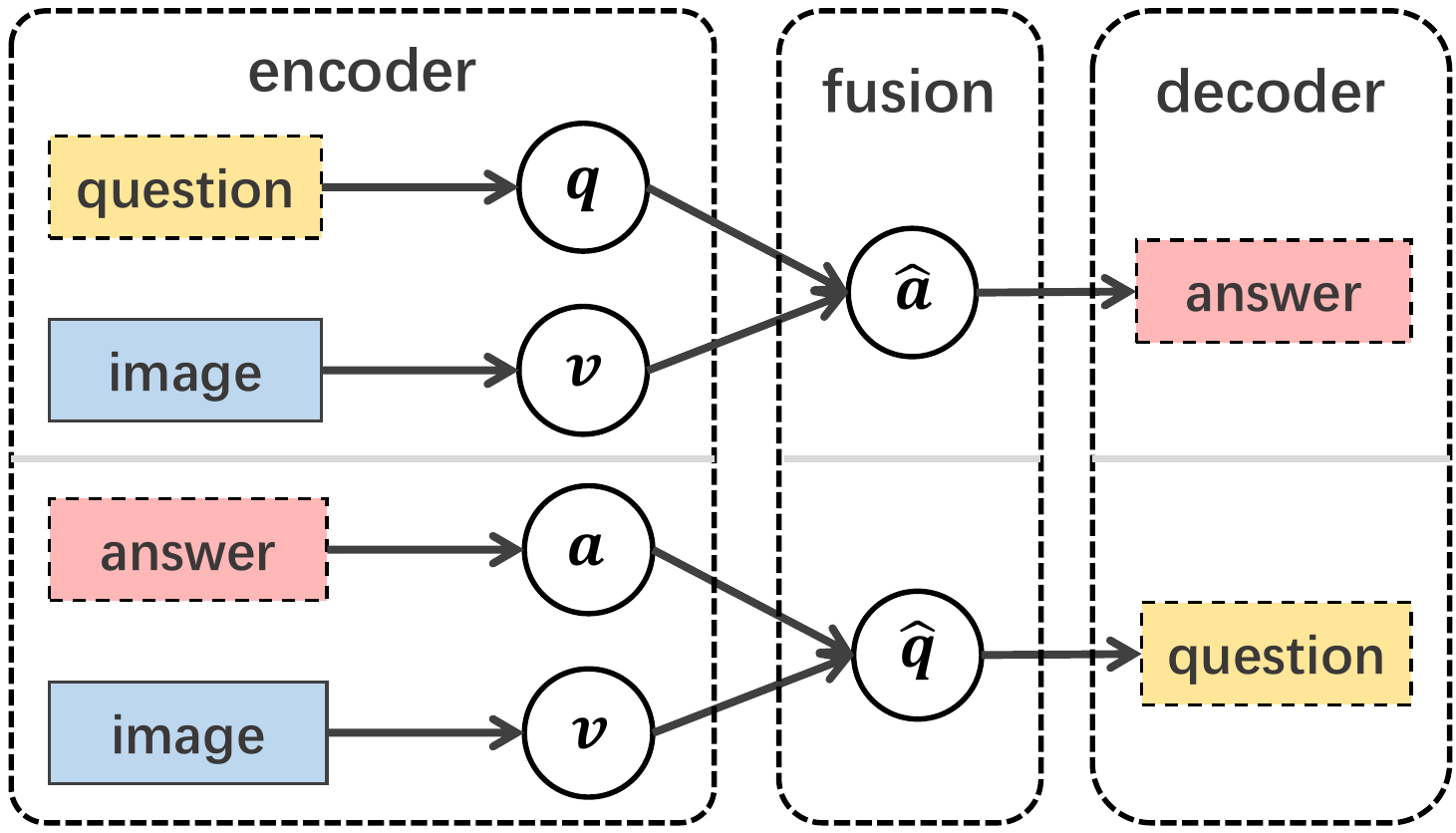}
		\end{center}
		\caption{Problem solving schemes of VQA~(top) and VQG~(bottom), both of which utilize the $\langle$\emph{encoder-fusion-decoder}$\rangle$ pipeline with $Q$ and $A$ in inverse order. $\mathbf{v}$, $\mathbf{q}$ and $\mathbf{a}$ respectively denote the encoded features of input image, question, and answer, while $\hat{\mathbf{a}}$ and $\hat{\mathbf{q}}$ represent the predicted answer/question features.}
		\label{fig:abstract}
	\end{figure}
	
	In previous works, VQA and VQG are studied separately. 
	As shown in Figure~\ref{fig:abstract}, the VQA model encodes the question sentence as an embedding $\mathbf{q}$, then associates $\mathbf{q}$ with the image feature $\mathbf{v}$ to infer the answer embedding $\hat{\mathbf{a}}$, which is decoded as the distribution over the answer vocabulary. Different from VQA, VQG has not got a standard problem setting. In this work, we formulate VQG as generating a question with an image and answer given. 
	where the VQG model merges the answer embedding $\mathbf{a}$ and the image feature $\mathbf{v}$ to get the question embedding $\hat{\mathbf{q}}$, and then generates a question sentence with recurrent neural network~(RNN).
	We can see that these two tasks are intrinsically correlated, \ie sharing visual input and taking encoder-fusion-decoder pipeline with Q and A in reverse order. Thus, we refer them as ``Dual'' tasks.

	Duality reflects the intrinsic complementary relation between question answering and generation. 
	Intuitively, learning to answer questions may boost the question generation and vice versa, as both of them require similar abilities: image recognition, question reasoning, cross-modal information association, etc. Thus, we argue that jointly learning through these two tasks can utilize the training data in a more efficient way, and can bring mutual improvements to both VQA and VQG.
	Therefore, we formulate the dual training of VQA and VQG as learning an invertible cross-modality fusion model that can infer Q or A when given the counterpart based on the given image.

	From this perspective, we derive an invertible \emph{Dual Mutan} fusion module, based on the state-of-the-art VQA model Mutan~\cite{ben2017mutan}. The module can complete the feature inference in a bidirectional manner, \ie it can infer the answer embeddings from image+question and infer the question embeddings from image+answer. Furthermore, by sharing the visual features as well as the encoder and decoder of the question and answer, VQG and VQA models can be viewed as the inverse form of each other with parameters shared. When jointly training on the two tasks, the invertibility brought by our parameter sharing schemes can help to regularize the training process, and multiple training tasks will help the model learn more general representations.

	\textbf{Contribution}: 
	This work is the first attempt to consider VQG and VQA as dual tasks and formulate them into a unified framework called Invertible Question Answering Network~(iQAN). The model is jointly trained with VQA and VQG tasks and can be deployed for either task in the testing stage.
	In the iQAN, a novel parameter sharing scheme and duality regularization are proposed to explicitly leverage the complementary relations between questions and answers. 
	Evaluated on VQA2 and CLEVR datasets, our proposed model achieves better results on VQA task. Experimental results show that our framework can also generalize to some other VQA models and continuously improves their performance. Besides, we propose a method to utilize the VQG model to augment questions with ground-truth answers given, which could employ the cheaply-labeled answers to boost the model training.

	\section{Related Work}
	
	
	\textbf{Visual Question Answering~(VQA)} is one of the most popular cross-discipline tasks aiming at understanding the question and image, and then providing the correct answer. Malinowski~\etal propose an encoder-decoder framework to merge the visual and textual information for answering prediction~\cite{malinowski2015ask}. Shih~\etal introduce visual attention mechanism to highlight the image regions relevant to answering the question~\cite{shih2016look}. Lu~\etal further apply attention to the language model, called co-attention, to jointly reason about images and questions~\cite{lu2016hierarchical}. 
	Apart from proposing new frameworks, some focus on designing effective multimodal feature fusion schemes~\cite{fukui2016multimodal,kim2017hadamard}. The bilinear model Mutan proposed by Ben-younes~\etal is the state-of-the-art method to model interactions between two modalities~\cite{ben2017mutan}. 
	Additionally, several benchmark dataset are proposed to facilitate the VQA research~\cite{malinowski2014towards}. VQA2 is the most popular open-ended Q-A dataset~\cite{balanced_vqa2} with real images. Johnson~\etal propose CLEVR dataset with rendered images and automatically-generated questions to mitigate answer biases and diagnose the reasoning ability of VQA models. We will evaluate our method on these two datasets.  

	\textbf{Visual Question Generation.} Question generation from text corpus has been investigated for years in natural language processing~\cite{ali2010automation,kalady2010natural,serban2016generating}. Recently, it has been introduced to computer vision to generate image-related questions. Mora~\etal propose a CNN-LSTM model to directly generate image-related questions and corresponding answers~\cite{moratowards}. Mostafazadeh~\etal collect the first VQG dataset, where each image is annotated with several questions~\cite{mostafazadeh2016generating}. Zhang~\etal propose a model to automatically generate visually grounded questions~\cite{zhang2016automatic}, which uses Densecap~\cite{densecap} to generate region captions as extra information to guide the question generation. Jain~\etal combine the variational autoencoder and LSTM to generate diverse questions~\cite{jain2017creativity}. Different from the existing works to generate question solely based on images, we provide an annotated answer as an additional cue. Therefore, VQG can be modeled as a two modalities fusion problem like VQA. 
	
	\begin{figure*}[t]
		\begin{center}
			\includegraphics[width=\linewidth]{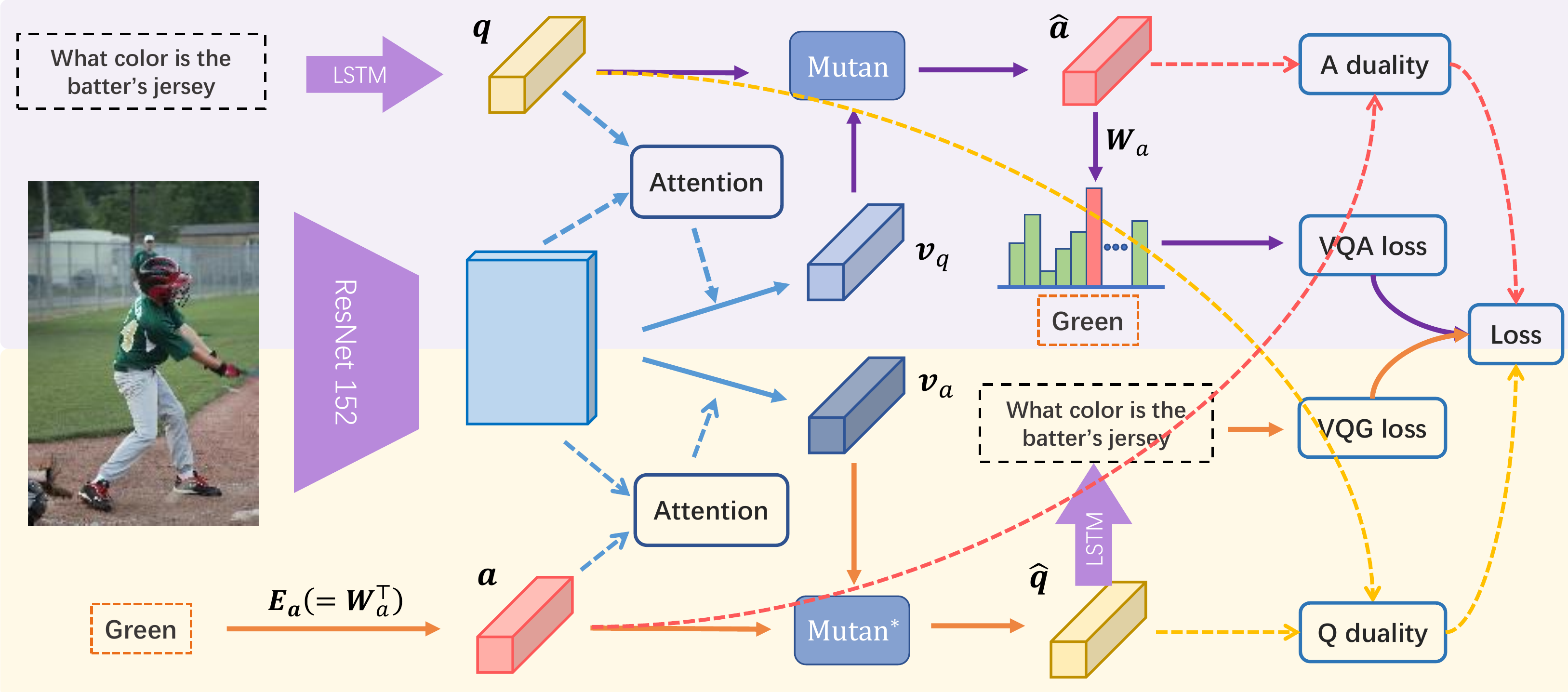}
		\end{center}
		\caption{Overview of Invertible Question Answering Network~(iQAN), which consists two parts, VQA and VQG. The upper part is Mutan VQA component~\cite{ben2017mutan}, and the lower part is its dual VQG model. Input questions and answers are encoded respectively by an \textbf{LSTM} and a lookup table~$\mathbf{E}_a$ into fixed-length features. With attention and Mutan fusion module, predicted features are obtained. The predict features are used for obtaining output~(by \textbf{LSTM} and $\mathbf{W}_a$ for questions and answers respectively). \textbf{A duality} and \textbf{Q duality} are duality regularizers to constrain the similarity between the answer and question representations in both models. 
		Two components share the \textbf{LSTM}, \textbf{Mutan} and \textbf{Attention} Modules. \textbf{Mutan$^*$} denotes the dual form of \textbf{Mutan}. $\mathbf{E}_a$ also shares parameters with $\mathbf{W}_a$.}
		\label{fig:DT-CNN}
	\end{figure*}
	
	\textbf{Dual Learning.} Utilizing cycle consistency to regularize the training process has a long history. It has been use as a standard trick for years in visual tracking to enforce forward-backward consistency~\cite{sundaram2010dense}.
	He~\etal first formulate the idea as \emph{Dual Learning} in machine translation area~\cite{he2016dual}, which uses A-to-B and B-to-A translation models to form a closed translation loop~(A-B-A and B-A-B), and lets them teach each other through a reinforcement learning process. Tang~\etal introduce the idea to QA area, where question generation is modeled as dual task of QA, and leverage the probabilistic correlation between QA and QG to guide the training~\cite{tang2017question}. Zhu~\etal employ the idea in computer vision and propose the image translation model CycleGAN~\cite{zhu2017unpaired}. However, till now there is no existing work that utilizes dual learning on VQA. Hence, our work is the first attempt to model the VQA and VQG as dual tasks and leverage the complementary relations between the two tasks.

	\section{Invertible Question Answering Network (iQAN)}
	
	In this section, we present the dual learning framework of the VQA and VQG, Invertible Question Answering Network~(iQAN). The overview of our proposed iQAN is shown in Figure~\ref{fig:DT-CNN}, which consists of two components, VQA component~(top) and VQG component~(bottom).
	
	In the VQA component, given a question $q$, LSTM is used for obtaining the embedded features $\mathbf{q}\in \mathbb{R}^{d_q}$, and CNN is used transformed the input image $v$ into an feature map. A Mutan-based attention module is used to generate a question-related visual feature $\mathbf{v}_q\in \mathbb{R}^{d_v}$ from the image and the question. Then a Mutan fusion module is used for obtaining the answer features $\hat{\mathbf{a}}\in \mathbb{R}^{d_a}$ from $\mathbf{v}_q$ and $\mathbf{q}$. Finally, a linear classifier~$\mathbf{W}_a$ is used to predict the answer for VQA. 
	
	In the VQG component, given an answer, a lookup table $\mathbf{E}_a$ is used for obtaining the embedded feature $\mathbf{a}\in \mathbb{R}^{d_a}$. CNN with attention module is used for obtaining the visual feature $\mathbf{v}_a\in \mathbb{R}^{d_v}$ from the input image and the answer feature $\mathbf{a}$. Then the Mutan in the dual form, which shares parameters with VQA Mutan but in a different structure, is used for obtaining the predicted question features $\hat{\mathbf{q}}\in \mathbb{R}^{d_q}$. Finally, an LSTM-based decoder is employed to translate $\hat{\mathbf{q}}$ to the question sentence.
	
	We will formulate the VQA and VQG components as inverse process to each other by introducing a novel parameter sharing scheme and the duality regularizer. Therefore, we could jointly train one model with two tasks to leverage the dependencies of questions and answer in a bidirectional way. In addition, the invertibility of the model could serve as a regular term to guide the training process.

	
	
	\subsection{The VQA component}\label{sec:vqa}
	The VQA component of our proposed iQAN is based on the state-of-the-art Mutan VQA model. We will briefly review the core part, Mutan fusion module, which takes an image feature $\mathbf{v}_q$ and a question feature $\mathbf{q}$ as input, and predicts the answer feature $\hat{\mathbf{a}}$.
	
	
	\subsubsection{Review on MUTAN fusion module}
	
	Since language and visual representations are in different modalities, the issue of merging visual and linguistic features is crucial in VQA. Bilinear models are recent powerful solutions to the multimodal fusion problem, which encode bilinear interactions between $\mathbf{q}$ and $\mathbf{v}_q$ as follows:
	\begin{equation}
	\hat{\mathbf{a}}=\left( \mathcal{T} \times_1 \mathbf{q}\right) \times_2 \mathbf{v}_q
	\label{eq:bilinear}
	\end{equation}
	where the tensor $\mathcal{T}\in \mathbb{R}^{d_q\times d_v\times d_a}$ denotes the fully-parametrized operator for answer feature inference, and $\times_i$ denotes the \emph{mode-i} product between a tensor and a matrix:
	\begin{equation}
	\left( \mathcal{T} \times_i \mathbf{U}\right)[d_1,...d_{i-1}, j,d_{i+1}...d_N] = \sum^{D_i}_{d_i=1} \mathcal{T}[d_1 ... d_N] \mathbf{U}[d_i,j]
	\end{equation}
	
	To reduce the complexity of the full tensor $\mathcal{T}$, Tucker decomposition~\cite{ben2017mutan}  is introduced as an effective way to factorize $\mathcal{T}$ as a tensor product between factor matrices $\mathbf{W}_q$, $\mathbf{W}_v$ and $\mathbf{W}_a$, and a \emph{core tensor} $\mathcal{T}_c$:
	\begin{equation}
	\mathcal{T}=\left( \left(\mathcal{T}_c \times_1 \mathbf{W}_q \right) \times_2 \mathbf{W}_v \right) \times_3 \mathbf{W}_a
	\label{eq:tucker}
	\end{equation}  
	with $\mathbf{W}_q \in \mathbb{R}^{t_q \times d_q}$,  $\mathbf{W}_v \in \mathbb{R}^{t_v \times d_v}$ and $\mathbf{W}_a \in \mathbb{R}^{t_a\times d_a}$, and $\mathcal{T}_c \in \mathbb{R}^{t_q\times t_v\times t_a}$. Consequently, we can rewrite Eq.~\ref{eq:bilinear} as:
	\begin{equation}
	\hat{\mathbf{a}}=\left(\left( \mathcal{T}_c \times_1 \left( \mathbf{W}_q \mathbf{q}\right)\right)
	\times_2 \left(\mathbf{W}_v \mathbf{v}_q\right)\right)
	\times_3 \mathbf{W}_a
	\label{eq:mutan_bilinear}
	\end{equation}
	where matrices $\mathbf{W}_q$ and $\mathbf{W}_v$ transform the question features $\mathbf{q}$ and image features $\mathbf{v}_q$ into dimensions $t_q$ and $t_v$ respectively. The squeezed bilinear core $\mathcal{T}_c$ models the interactions among the transformed features and projects them to the answer space of size $t_a$, which is used to infer the per-class score by $\mathbf{W}_a$.
	
	If we define $\tilde{\mathbf{q}}=\mathbf{W}_q \mathbf{q}\in\mathbb{R}^{t_q}$ and $\tilde{\mathbf{v}}_q=\mathbf{W}_v \mathbf{v}_q\in\mathbb{R}^{t_v}$, then we have:
	\begin{equation}
	\tilde{\mathbf{a}}=\left(\mathcal{T}_c \times_1 \tilde{\mathbf{q}}\right)\times_2 \tilde{\mathbf{v}}_q\in \mathbb{R}^{t_a}
	\label{eq:mutan_bilinear_2}
	\end{equation}
	Thus, $\tilde{\mathbf{a}}$ can be viewed as the \emph{answer feature} where $ \hat{\mathbf{a}} = \tilde{\mathbf{a}}^\top \times \mathbf{W}_a$. 
	
	To balance the complexity and expressivity of the interaction modeling, the low rank assumption is introduced, and $\mathcal{T}_c\left[:,:,k\right]$ can be expressed as a sum of $R$ rank one matrices:
	\begin{equation}
	\mathcal{T}_c\left[:,:,k\right]=\sum^{R}_{r=1} \mathbf{m}_r^k\otimes {\mathbf{n}_r^k}^\top
	\label{eq:R_matrices_sum}
	\end{equation}
	with $\mathbf{m_r^k}\in \mathbb{R}^{t_q}$,  $\mathbf{n_r^k}\in \mathbb{R}^{t_v}$ and $\otimes$ denoting the outer product. Then each element $\tilde{\mathbf{a}}\left[k\right]$ of $\tilde{\mathbf{a}}$, $k\in \{ 1, \ldots, t_a \}$, can be written as:
	\begin{equation}
	\tilde{\mathbf{a}}\left[k\right]=\sum^{R}_{r=1} \left(\tilde{\mathbf{q}}^\top\mathbf{m}_r^k  \right) \left(\tilde{\mathbf{v}}_q^\top\mathbf{n}_r^k  \right)
	\label{eq:slice_a_2}
	\end{equation}
	We can define $R$ matrices $\mathbf{M}_r\in \mathbb{R}^{t_q\times t_a}$ and $\mathbf{N}_r\in \mathbb{R}^{t_v\times t_a}$ such that $\mathbf{M}_r\left[:,k\right] = \mathbf{m}_r^k$ and $\mathbf{N}_r\left[:,k\right] = \mathbf{n}_r^k$. Therefore, with sparsity constraints, Eq.~\ref{eq:mutan_bilinear_2} is further simplified as:
	\begin{equation}
	\tilde{\mathbf{a}}=\sum^{R}_{r=1} \left(\tilde{\mathbf{q}}^\top\mathbf{M}_r  \right) \odot \left(\tilde{\mathbf{v}}_q^\top\mathbf{N}_r  \right)
	\label{eq:slice_a_3}
	\end{equation}
	where $\odot$ denotes the element-wise product. With Mutan, low computational complexity and strong expressivity of the model are both obtained.

	\subsection{The VQG component}\label{sec:vqg}
	
	The VQG component of our proposed iQAN is formulated as generating a question~(word sequence) given an image and an answer label. 
	
	During training, our target is to learn a model such that the generated question $\hat{q}$ similar to the referenced one $q^*$. The generation of each word of the question can be written as:
	\begin{equation}
	\hat{w}_t = \arg\max_{w\in \mathbb{W}} p \left(w\big\vert v,w_0, ..., w_{t-1}\right)
	\label{eq:vqg}
	\end{equation}
	where $\mathbb{W}$ denotes the word vocabulary. $\hat{w}_t$ is the predicted word at $t$ step. $w_i$ represents the $i$-th ground-truth word. Beam search will be used during inference. 
	
	VQG shares the visual CNN with VQA part. The answer feature $\mathbf{a} \in \mathbb{R}^{d_a}$ is directly retrieved from the answer embedding table~$\mathbf{E}_a$. MUTAN is also utilized for visual attention module and visual \& answer representations fusion at VQG. Similar to Eq.~\ref{eq:slice_a_3}, the inference of question features $\tilde{\mathbf{q}}$ can be formulated as:
	\begin{equation}
	\tilde{\mathbf{q}}=\sum^{R}_{r=1} \left(\tilde{\mathbf{a}}^\top\mathbf{M}^\prime_r  \right) \odot \left(\tilde{\mathbf{v}}_a^\top\mathbf{N}_r^\prime  \right)
	\label{eq:infer_q}
	\end{equation}
	with $\tilde{\mathbf{a}} = \mathbf{W}_a \mathbf{a}\in \mathbb{R}^{t_a}$ and $\tilde{\mathbf{v}}_a=\mathbf{W}_v \mathbf{v}_a\in\mathbb{R}^{t_v}$.  $\mathbf{M}^\prime_r$ and $\mathbf{N}^\prime_r$ are defined as Eq.~\ref{eq:slice_a_3}.
	Finally, the predicted question features $\tilde{\mathbf{q}}$ is fed into an RNN-based model to generate the predicted question. 
	
	From the formulation in (\ref{eq:slice_a_3}) and (\ref{eq:infer_q}), the VQG Mutan could be viewed as the conjugate form of the VQA Mutan. In the next section, we will introduce our attempt to investigate the connection between the two Mutan modules.

	\begin{figure}[t]
		\begin{center}
			\includegraphics[width=\linewidth]{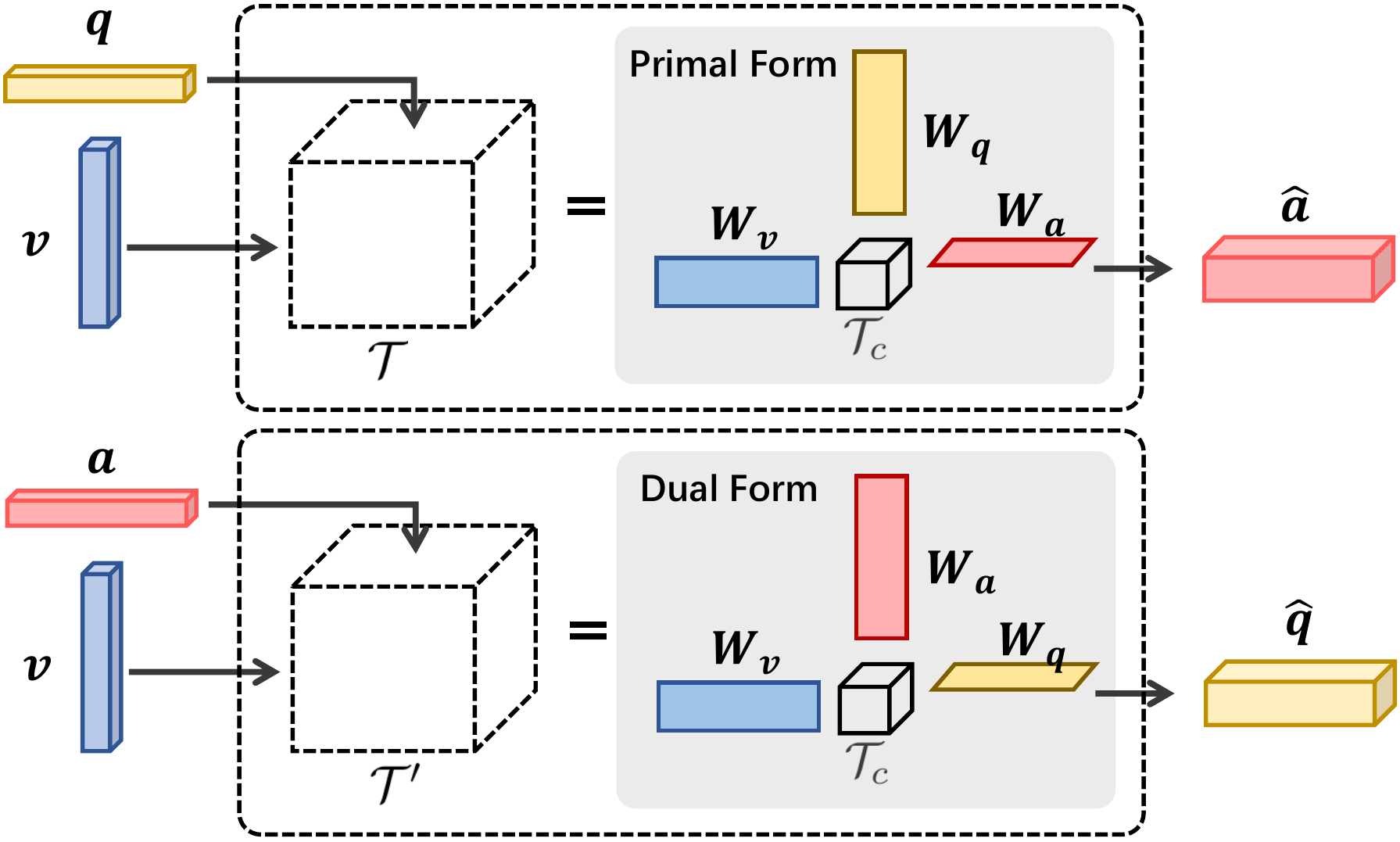}
		\end{center}
		\caption{The Dual Mutan in the primal form~(for VQA) and dual form~(for VQG). The two forms share one parameter set: the core tensor~$\mathcal{T}^c$, projection matrices of images, questions, and answers. In our experiment, $\mathbf{W}_a$ at the top part and $\mathbf{W}_q$ at the bottom part are merged with decoders.}
		\label{fig:dual_mutan}
	\end{figure}
	
	\subsection{Dual MUTAN}
	To leverage the duality of questions and answers, we derive a \emph{Dual Mutan} from the original Mutan to finish the primal~(question-to-answer) and its dual~(answer-to-question) inference on the feature level  with one kernel. 
	
	First we rewrite Eq.~\ref{eq:mutan_bilinear_2} and its dual form:
	\begin{equation}
	\begin{aligned}
	&\tilde{\mathbf{a}}^*=\left( \mathcal{T}_c \times_1 \tilde{\mathbf{q}}\right) \times_2 \tilde{\mathbf{v}} \\
	&\tilde{\mathbf{q}}^*=\left( \mathcal{T}_c^\prime \times_1 \tilde{\mathbf{a}}\right) \times_2 \tilde{\mathbf{v}}^\prime
	\end{aligned}
	\end{equation}
	where $\mathcal{T}_c\in \mathbb{R}^{t_q \times t_v \times t_a}$, $\mathcal{T}_c^\prime \in \mathbb{R}^{t_a \times t_v \times t_q}$, $ \tilde{\mathbf{q}} =  \mathbf{W}_q \mathbf{q}$, $ \tilde{\mathbf{a}} = \mathbf{W}_a \mathbf{a}$, 
	$ \tilde{\mathbf{v}} = \mathbf{W}_v \mathbf{v}$, and $ \tilde{\mathbf{v}}^\prime = \mathbf{W}_v^\prime \mathbf{v}$. For simplicity, it is assumed that both VQA and VQG adopt $\mathbf{v}$ as visual input, which can be replaced by the post-attention feature $\mathbf{v}_a$ or $\mathbf{v}_q$. 
	Noticing both $\mathcal{T}_c$ and $\mathcal{T}_c^\prime$ model the interactions among the image, question, and answer embeddings, but with different dimension arrangement, we assume the relationship between $\mathcal{T}_c^\prime$ and $\mathcal{T}_c$ as follows:
	\begin{equation}
	\mathcal{T}_c^\prime[:, i, :] = \mathcal{T}_c^\top[:, i, :] 
	\end{equation}
	Additionally, the transform matrices for visual information $\mathbf{W}_v^\prime$ and $\mathbf{W}_v$ can also be shared. Therefore, we can unify the question and answer embedding inference with single three-way operator~$\mathcal{T}_c$:
	\begin{equation}
	\begin{aligned}
	\tilde{\mathbf{a}}^*=\left( \mathcal{T}_c \times_1 \tilde{\mathbf{q}}\right) \times_2 \tilde{\mathbf{v}} \\
	\tilde{\mathbf{q}}^*=\left( \mathcal{T}_c \times_3 \tilde{\mathbf{a}}\right) \times_2 \tilde{\mathbf{v}}
	\end{aligned}
	\label{eq:share_T}
	\end{equation}
	Furthermore, since $\mathcal{T}_c[:, i, :]$ represents the correlation between the re-parameterized question and answer embeddings, considering the duality of Q and A, we could assume the the following for $\mathcal{T}_c[:, i, :]$: 
	\begin{equation}
	\left\{
	\begin{aligned}
	&t_a = t_q = t \\
	&\mathcal{T}_c[:, i, :] = \mathcal{T}_c^\top[:, i, :], \ i \in \left[1, t_v\right]
	\end{aligned}\right.
	\end{equation}
	Correspondingly,  Eq.~\ref{eq:share_T} could be written as:
	\begin{equation}
	\begin{aligned}
	\tilde{\mathbf{a}}^*=\left( \mathcal{T}_c \times_1 \tilde{\mathbf{q}}\right) \times_2 \tilde{\mathbf{v}} \\
	\tilde{\mathbf{q}}^*=\left( \mathcal{T}_c \times_1 \tilde{\mathbf{a}}\right) \times_2 \tilde{\mathbf{v}}
	\end{aligned}
	\end{equation}
	That is to say, we could infer $\tilde{\mathbf{a}}$ or $\tilde{\mathbf{q}}$ by just alternating the \emph{mode-1} input of the kernel. 
	
	By introducing the sparsity constraint like Eq.~\ref{eq:slice_a_3}, the inference of answer and question features $\tilde{\mathbf{a}}^*$ and $\tilde{\mathbf{q}}^*$ can be reformulated as:
	\begin{equation}
	\begin{aligned}
	\tilde{\mathbf{a}}^*=\sum^{R}_{r=1} \left(\tilde{\mathbf{q}}^\top\mathbf{M}_r  \right) \odot \left(\tilde{\mathbf{v}}^\top\mathbf{N}_r  \right) \\
	\tilde{\mathbf{q}}^*=\sum^{R}_{r=1} \left(\tilde{\mathbf{a}}^\top\mathbf{M}_r  \right) \odot \left(\tilde{\mathbf{v}}^\top\mathbf{N}_r  \right)
	\end{aligned}
	\end{equation}
	And the target answer and question embeddings are provided by:
	\begin{equation}
	\begin{aligned}
	\hat{\mathbf{a}} =  \tilde{\mathbf{a}}^{*\top} \times \mathbf{W}_a \\
	\hat{\mathbf{q}} = \tilde{\mathbf{q}}^{*\top} \times \mathbf{W}_q 
	\label{eq:final_pred}
	\end{aligned}
	\end{equation} 
	As shown in Fig.~\ref{fig:dual_mutan}, we unify the two Mutan modules by sharing parameters $\mathbf{W_a}$, $\mathbf{W}_q$, $\mathbf{W}_v$, and $\mathcal{T}_c$. And we call this invertible module \emph{Dual Mutan}. 
	
	Furthermore, when the decoder after the dual Mutan are considered, the predicted answer embedding $\hat{\mathbf{a}}$  will be fed into another linear transform layer to get the per-class score, and the question embedding $\hat{\mathbf{q}}$ will be decoded by LSTM, both of which have linear transforms afterwards. So the linear transforms in Eq.~\ref{eq:final_pred} can be skipped for efficiency. And we can directly use $\tilde{\mathbf{a}}^{*}$ and $\tilde{\mathbf{q}}^{*}$ as the predicted features to feed into decoders.

	\subsection{Parameter Sharing for Q and A encoding and decoding.}
	Considering the duality of VQA and VQG, the encoder and decoder of Q/A can be viewed as reverse transformation to each other. Hence, we could employ these properties to propose corresponding weight sharing scheme.
	
	For input answers in the VQG component, the input answer is embedded into features $\mathbf{a}$ by the matrix $\mathbf{E}_a$, which stores the embeddings of each answer. For the answer generation in the VQA component, the predicted feature $\hat{\mathbf{a}}$ is decoded for obtaining the answer through a linear classifier $\mathbf{W}_a$, which can be regarded as a set of per-class templates for the feature matching. Thus, we can directly share the weights of $\mathbf{E}_a$ and $\mathbf{W}_a$, where $\mathbf{E}_a$ is required to be the transposition  of  $\mathbf{W}_a$.
	
	For input questions in the VQA component, LSTM is applied to encode the question sentence into a fixed-size feature vector $\mathbf{q}$. For the question generation in the VQG component, LSTM is also applied to decode the vector back to a word sequence step-by-step. We also share the parameters of the two LSTMs. 
	Experimental results show that sharing weights of two LSTMs will not deteriorate the final result but requires fewer parameters. 

	\subsection{Duality Regularizer}
	With Dual Mutan, we have reformulated the feature fusion part of VQA and VQG~($\phi$ and $\phi^*$)  as the inverse process to each other. $\phi$ and $\phi^*$ are expected to form a closed cycle. 
	Consequently, given a question/answer pair~$(\mathbf{q},\mathbf{a})$, the predicted answer/question representations are expected to have the following form:
	\begin{equation}
	\mathbf{a}\approx \hat{\mathbf{a}}=\phi(\mathbf{q}, \mathbf{v}) \ \text{and}\ \mathbf{q}\approx \hat{\mathbf{q}}=\phi^*(\mathbf{a}, \mathbf{v}). 
	\end{equation}
	To leverage the property above, we propose the Duality Regularizer, $\text{smooth}_{L1}\left(\hat{\mathbf{q}} - \mathbf{q}\right)$ and $\text{smooth}_{L1}\left(\hat{\mathbf{a}} - \mathbf{a}\right)$, 
	where the loss function $\text{smooth}_{L1}$ is defined as:
	\begin{equation}
	\text{smooth}_{L1}(x) = \left\{ 
	\begin{aligned}
	& 0.5 * x^2, \ &\text{if}\ |x| < 1 \\
	& |x| - 0.5,\  &\text{otherwise}
	\end{aligned}\right. \label{eq:smoothL1}
	\end{equation}
	By minimizing Q/A duality loss, primal and dual question/answer representations are unified, and VQG and VQA are linked with each other. Moreover, the Duality Regularizer can be viewed as a way to provide soft targets for the question/answer feature learning.

	\subsection{Dual Training}
	
	With our proposed weight sharing schema~(Dual Mutan and Sharing De-/Encoder), our VQA and VQG models can be reconstructed to inverse process of each other with sharing parameters. Hence, joint training on VQG and VQA tasks introduces the \emph{invertibility} of the model as an additional regular term to regulate the training process. 
	The overall training loss including our proposed Q/A duality is as below:
	\begin{equation}
	\begin{aligned}
	Loss = & L_{(vqa)}\left(a, a^*\right) + L_{(vqg)}\left(q, q^*\right)  \\
	& + \text{smooth}_{L1}\left(\mathbf{q} - \hat{\mathbf{q}}\right) + \text{smooth}_{L1}\left(\mathbf{a} - \hat{\mathbf{a}}\right)
	\label{eq:loss}
	\end{aligned}
	\end{equation}
	where $L_{(vqa)}\left(a, a^*\right)$ and $L_{(vqg)}\left(q, q^*\right)$ adopt the multinomial classification loss~\cite{ben2017mutan} and sequence generation loss~\cite{vinyals2015show} as the unary loss for VQA and VQG components respectively.
	
	Additionally, as every operation is differentiable, the entire model can be trained in an end-to-end manner. In the next section, we will show that our dual training strategy could bring significant improvement for both VQA and VQG models. 
	
	\begin{table*}[t]
		\renewcommand{\arraystretch}{1.1}
		\setlength{\tabcolsep}{4pt}
		\small
		\begin{center}
			\begin{tabularx}{\linewidth}{c|ccc|cc|cc|cccc |c}
				\hline
				\multirow{2}{*}{model} & \multirow{2}{*}{\specialcell{Dual  Mutan}} & 
				\multirow{2}{*}{\specialcell{Duality \\ Regularizer}}  & \multirow{2}{*}{\specialcell{Sharing \\ De- \& Encoder} } & 
				\multicolumn{2}{c}{\specialcell{VQA2 Subset\\ \cite{balanced_vqa2}}} \vline& 
				\multicolumn{2}{c}{\specialcell{VQA2 Full\\ \cite{balanced_vqa2}}} \vline&
				\multicolumn{5}{c}{\specialcell{CLEVR\\ \cite{johnson2016clevr}}}\\
				&&&&acc@1&acc@5&acc@1&acc@5& size & material & shape & color & \textbf{Overall}\\
				\hline
				\hline
				1 &   - 			& - 			& - 			& 50.72 & 78.56 & 54.83 & 87.68 & 86.76 & 88.25 & 82.26 & 76.86  &  83.74\\
				2 &    \checkmark 	& - 			& - 			& 50.99 & 78.71 & 54.78 & 87.66 & 87.29 & 88.10 & 82.82 & 77.62 & 84.13 \\
				3 &    \checkmark 	& - 			& \checkmark 	& 51.23 & 78.80 & 54.70 & \textbf{88.07} & 87.42 & 88.81 & 82.84 & 77.73 &  84.40\\
				4 &    \checkmark 	& \checkmark 	& - 			& 51.38 & \textbf{78.96} & 54.39 & 87.92 & 87.75 & 88.48 & \textbf{84.28} & 77.97 &  84.78 \\
				5 &    \checkmark 	& \checkmark 	& \checkmark 	& \textbf{51.49} & 78.93 & \textbf{54.97} & 87.77 & \textbf{87.75} & \textbf{88.91} & 84.08 & \textbf{78.86}  & \textbf{85.07}\\
				\hline
			\end{tabularx}
		\end{center}
		\caption{Ablation study of different settings. \textbf{Dual Mutan}: our proposed sharing Mutan scheme. \textbf{Duality Regularizer}: an additional regular term defined in Eq.~(\ref{eq:smoothL1}) and (\ref{eq:loss}) to guarantee the similarity of dual pairs~($\mathbf{q}\approx \hat{\mathbf{q}}$ and $\mathbf{a}\approx \hat{\mathbf{a}}$). \textbf{Sharing De- \& Encoder}: parameter sharing scheme for decoders and encoders of Q and A. Model 1 is the baseline model with separated VQA and VQG models.  Additionally, the per-question-type top-1 accuracies on CLEVR are also listed.}
		\label{tab:component}
	\end{table*}

	\section{Experiments}
	Model implementation details, data preparation and experiment results will be shown in this section. Qualitative results will be shown in the supplementary materials. 
	
	\subsection{Implementation Details}
	Our iQAN is based on PyTorch implementation of Mutan VQA~\cite{ben2017mutan}. We directly use the ImageNet-pretrained ResNet-152~\cite{resnet} provided by PyTorch as our base model and keep this part fixed. All images are resized to $448\times448$, and the size of feature maps is $14\times14$. Newly introduced parameters are randomly initialized. Adam~\cite{adam} with fixed learning rate 0.0001 is used to update the parameters. The training batch size is 512.\footnote{Batch size will influence the model performance. To be fair, we use 512 for all experiments.} All models are trained for 50 epochs, and the best validation results are used as final results. 
	
	\subsection{Data Preparation}
	We evaluate the proposed method on two large-scale VQA datasets, VQA2~\cite{balanced_vqa2} and CLEVR~\cite{johnson2016clevr}, both of which provide images and labeled $\langle$Q,A$\rangle$ pairs. However, these two datasets contain some of the questions with non-informative answers as ~\emph{yes/no} or \emph{number}. It is nearly impossible for a model to generate expected questions from an answer like \emph{yes}. 
	Therefore, we preprocess the data to filter out some question-answer pairs for both the VQA2 and the CLEVR to fairly explore the duality of Q and A:  For VQA2, we only select the questions whose annotated question type starts with ``what", ``where" or ``who". For CLEVR, the questions starting with ``what" and whose answer is not \emph{number} are selected. 
	Additionally, for VQA2, we fixed the answer vocabulary size to 2000 most frequent answers as in~\cite{ben2017mutan}. $\langle$Q,A$\rangle$ pairs whose answer is not in the vocabulary will be removed. 
	Detailed statistics of filtered VQA2 and CLEVR are shown in Table~\ref{tab:dataset}.

	\begin{table}[t]
		\renewcommand{\arraystretch}{1.2}
		\setlength{\tabcolsep}{4pt}
		\begin{center}
			\begin{tabularx}{0.92\linewidth}{l cc cc}
				\hline
				\multirow{2}{*}{Dataset} & \multicolumn{2}{c}{Train} & \multicolumn{2}{c}{Validation}\\
				& \#images & \#Q,A pairs & \#images & \#Q,A pairs \\
				\hline
				VQA2 & 163,550 & 68,434 & 78,047 & 33,645 \\
				CLEVR & 107,132 & 57,656 & 22,759 & 12,365 \\
				\hline
			\end{tabularx}
		\end{center}
		\caption{Statistics of filtered VQA2~\cite{balanced_vqa2} and CLEVR~\cite{johnson2016clevr}.}
		\label{tab:dataset}
	\end{table}
	
	\begin{table*}[t]
		\renewcommand{\arraystretch}{1.2}
		\setlength{\tabcolsep}{4pt}
		\small
		\begin{center}
			\begin{tabularx}{0.95\linewidth}{l | ccc | ccc | ccc | ccc}
				
				\hline
				\multirow{2}{*}{Model} & \multicolumn{3}{c}{iBOWIMG~\cite{zhou2015simple}}\vline& \multicolumn{3}{c}{MLB~\cite{kim2017hadamard}} \vline& \multicolumn{3}{c}{MUTAN\ +\ SkipThought}\vline& \multicolumn{3}{c}{MUTAN\ +\ LSTM} \\
				& Acc@1& Acc@5 & BLEU & Acc@1 & Acc@5 & BLEU & Acc@1 & Acc@5 & BLEU & Acc@1 & Acc@5 & BLEU\\\hline\hline
				Baseline & 42.05 & 72.79 & 55.23 & 50.23 & 77.64 & 55.35 & 50.72 & 78.56 & 54.15 & 49.91 &  77.47 & 54.17\\
				Dual Training& 43.44 & 74.27 & 55.36 & 50.83 & 78.12 & 55.60 & 51.49 & 78.93 & 54.83 & 50.78 & 78.16 & 54.89 \\\hline
				Gain & 1.39 & 1.48 & 0.13 & 0.60 & 0.48 & 0.25 & 0.77 & 0.37 & 0.68 & 0.87 & 0.69 & 0.72 \\
				\hline
			\end{tabularx}
		\end{center}
		\caption{Evaluation of Dual Training Scheme on different VQA models. \textbf{Acc@1} and \textbf{Acc@5} are the VQA metrics, while \textbf{BLEU} score is used to measure the question generation quality. \textbf{Baseline} models are separately-trained VQA and VQG. \textbf{Dual Training} is to employ our proposed parameter sharing schemes and Dual Regularizer. The Dual Training version is to train one model with two tasks while Baseline is to train two different models. \emph{SkipThought} and \emph{LSTM} denote two question encoders used in the model.  }
		\label{tab:multimodel}
	\end{table*}

	\subsection{Performance Metrics}
	VQA is commonly formulated as the multinomial classification problem while VQG is a sequence generation problem. Therefore, we use top-1 accuracy~(Acc@1) and top-5 accuracy~(Acc@5) to measure the quality of the predicted answers. Sentence-level BLEU score~\cite{papineni2002bleu} provided by NLTK~\cite{nltk} is employed to evaluate the generated questions~(with Method 4 smooth function). 
	
	\subsection{Component Analysis}
	We compare our proposed Dual Training scheme with the baseline Mutan model on three datasets, filtered VQA2, full VQA2 and filtered CLEVR. Table~\ref{tab:component} shows our investigation on different settings. Model 1 is the baseline model with separated VQA and VQG models. 
	
	First, we focus on the filtered VQA2 dataset. By comparing model 1 and 2, we can see that our proposed Dual Mutan can help to improve VQA but not significantly. Just as we discussed below, the derivation of Dual Mutan module is based on a lot of assumptions guaranteed by the duality regularizer and the encoder/decoder sharing. 
	When further adding these two components, the model performance is continuously improved, and the full model shown in 5 outperforms the baseline model by 0.77\% on top-1 accuracy, which is a significant improvement for VQA. 
	
	Furthermore, similar experiments are done on the full VQA2 dataset, but there is little improvement, while model 2$\sim$4 are even worse than the baseline model. That is mainly because generating expected questions from answers like \emph{yes} or \emph{no} is almost impossible, where the information provided by answers is too little for question generation. 
	Even worse, VQG loss will dominate the model training, which may deteriorate the VQA performance. 

	We also evaluate our proposed method on the CLEVR dataset, which is designed to diagnose the reasoning ability of VQA models. By comparing our full model and baseline model, we can see that our dual training scheme could help to improve the reasoning ability of the VQA model~(1.33\% gain on overall Acc@1). In addition, since VQA and VQG model are inverse form of each other, the dual training of VQG and VQA can be regarded as training a model to understand the question and then ask a similar one, so the model gets more training on reasoning ability. 
	
	\subsection{Dual Learning for Other VQA Models}
	
	Our proposed dual training mechanism can be viewed as reconstructing VQA model to finish VQG problem. By sharing parameters, the model is trained with two tasks. Although the dual training method is derived from Mutan, but the core idea can also be applied to other latest VQA models~\cite{zhou2015simple,kim2017hadamard}~(shown in Table~\ref{tab:multimodel}). 
	
	\noindent\textbf{iBOWIMG} is a simple baseline bag-of-word~(BOW) VQA model which simply concatenates image and question embeddings to predict the answer. Correspondingly, we implement a dual VQG model with similar feature fusion. Since there is no parameters for fusion part, Dual Training only requires decoder \& encoder sharing and duality regularizer. Experiment results show that jointly training VQG and VQA could bring mutual improvements to both, especially for VQA model~(1.39\% on Acc@1). However, the improvement for VQG is not significant, because iBOWIMG VQA uses BOW to encode questions while the VQG model uses LSTM to decode question features. Therefore, the compulsive similarity of predicted features for LSTM and BOW-encoded feature will be too strong as a regularizer. In addition, generating questions is hard and the baseline VQG performance is already high when comparing with other models, so there is little room for improvement. 
	
	\noindent\textbf{MLB} is another latest bilinear VQA model that can be viewed as the special case of Mutan which sets the core bilinear operator~$\mathcal{T}^c$ to identity. Therefore, the derived dual training scheme can be applied to MLB model directly and it can help VQG and VQA improve each other. 
	
	\noindent\textbf{Mutan + X}: The original Mutan model in \cite{ben2017mutan} utilizes the pretrained skip-thought model~\cite{kiros2015skip} as question encoder, so we change that to LSTM~(trained from scratch) to make it sharable with the decoder. For both versions, the dual training could consistently bring gains to VQA and VQG. Besides, by comparing two versions, we could find that the pretrained encoder performs better on VQA, which is a trick to improve VQA performance while hardly influencing VQG.
	
	By applying the dual training on three latest models, we can see that even though our proposed method is derived from Mutan, it can be generalized to other VQA models and bring concordant improvements. 

	
	\subsection{Augmenting VQA with VQG}
	\begin{table}[t]
		\renewcommand{\arraystretch}{1.2}
		\setlength{\tabcolsep}{4pt}
		\small
		\begin{center}
			\begin{tabularx}{\linewidth}{ll c c c}
				
				\hline
				Model & Dataset & Acc@1 & Acc@5 & BLEU\\
				\hline\hline
				Baseline 	& 0.5 $\langle$Q,A$\rangle$ 			& 46.68 	& 74.43 & 50.96\\
				DT 				& 0.5 $\langle$Q,A$\rangle$   			& 47.63 	& 75.42 & 53.23\\
				VQG+Baseline  		& 0.5 $\langle$Q,A$\rangle$ + 0.5 A  	& 47.51 	& 75.39 & 45.70\\
				VQG+DT 			& 0.5 $\langle$Q,A$\rangle$ + 0.5 A		& 47.99 	& 75.79 & 46.06\\
				VQG+DT+FT & 0.5 $\langle$Q,A$\rangle$ + 0.5 A		& \textbf{48.48} & \textbf{76.23} & \textbf{53.78}\\\hline\hline
				Baseline 	& 0.1 $\langle$Q,A$\rangle$ 			& 33.60 	& 61.04 & 47.26\\
				DT 			& 0.1 $\langle$Q,A$\rangle$ 			& 35.23 	& 62.77 & 48.45\\
				VQG+Baseline		& 0.1 $\langle$Q,A$\rangle$ + 0.9 A  	& 37.83 	& 64.86 & 44.90\\
				VQG+DT 			& 0.1 $\langle$Q,A$\rangle$ + 0.9 A		& 38.87 	& 66.02 & 44.92\\
				VQG+DT+FT 		& 0.1 $\langle$Q,A$\rangle$ + 0.9 A		& \textbf{39.95} & \textbf{66.67} & \textbf{49.40}\\
				\hline
			\end{tabularx}
		\end{center}
		\caption{Our investigation on augmenting $\langle$Q,A$\rangle$ pairs using VQG with A given. \textbf{Baseline} is separately trained VQA and VQG models. \textbf{DT} denotes the dual training. 0.1 and 0.5 denote the proportion of training data used as Set 1.}
		\label{tab:partial}
	\end{table}
	
	Since VQG could provide more questions given answers, besides as a dual task to train VQA model, VQG could also help to generate expensively-labeled questions from cheaply-labeled concepts~(answers) to produce more training data with little cost. So we propose two ways to employ the augmented data by VQG and evaluate them on filtered VQA2 dataset. In this section, the training set will be partitioned into two parts: one is with $\langle$Q,A$\rangle$ pairs~(Set 1); the other only contains answers~(Set 2).  Experiment results are shown in Table~\ref{tab:partial}. 
	
	\textbf{VQG+X}: We first train a VQG model~(with dual training) with on Set 1, and use it to generate questions given the answers in Set 2. Then we combine the Set 1 and augmented Set 2 as training data. \textbf{X} can be baseline or dual trained (DT) model. Results show that, compared to the model trained only on Set 1, such method will improve VQA but deteriorate VQG performance. It is mainly because the generated questions are not identical to the original ones, as one answer could correspond to several reasonable questions. So the generated questions may follow a different distribution. Hence, learning to generate questions from Set 2 will deteriorate the performance of VQG on the validation set. On the other hand, most of the generated questions can be answered by the given answer, which is the reason why they can serve as the augmented data to boost the VQA performance.
	
	\textbf{VQG+DT+FT}: Although Set 2 provides extra training data, the quality is not so good as Set 1. Therefore, a better way is to pretrain using dual training~(\textbf{DT}) for the model on Set 1 and augmented Set 2, and then finetune~(\textbf{FT}) the model on Set 1. From experiment results, we can see that our proposed data augmentation method significantly outperforms the vanilla dual-trained models~(VQG+DT+FT v.s. DT), our proposed method could successfully leverage the additional annotated answers to boost the model training.

	\section{Conclusion}
	We present the first attempt to consider visual question generation as a dual task of visual question answering. Correspondingly, we proposed a dual training scheme, iQAN, that is derived from Mutan VQA model but also applied to some other latest VQA models. Our proposed method could reconstruct VQA model to VQG and train a single model with two conjugate tasks. Experiments show that our dual trained model outperforms the baseline model on both VQA2 and CLEVR dataset, and it consistently brings gains to several latest VQA models. We further investigate the potential of using VQG to augment training data. Our proposed method is proved to be an effective way to leverage the cheaply-labeled answers to boost the VQA and VQG models. 
	
	\section{Acknowledgment}
	This work is supported by Hong Kong Ph.D.  Fellowship Scheme, SenseTime Group Limited and Microsoft Research Asia. We also thank Duyu Tang, Yeyun Gong, Zhao Yan, Junwei Bao and Lei Ji for helpful discussions.
	
	{\small
		\bibliographystyle{aaai}
		\bibliography{egbib}
	}

%
%
%
%
%
\onecolumn

\section{Supplementary Materials: Qualitative Results}

Question answering and question generation results on VQA2 validation set are shown in this section. Both VQA and VQG are jointly trained with our proposed dual training scheme, iQAN. 

\begin{figure*}[h]
	\begin{center}
		\includegraphics[width=\linewidth]{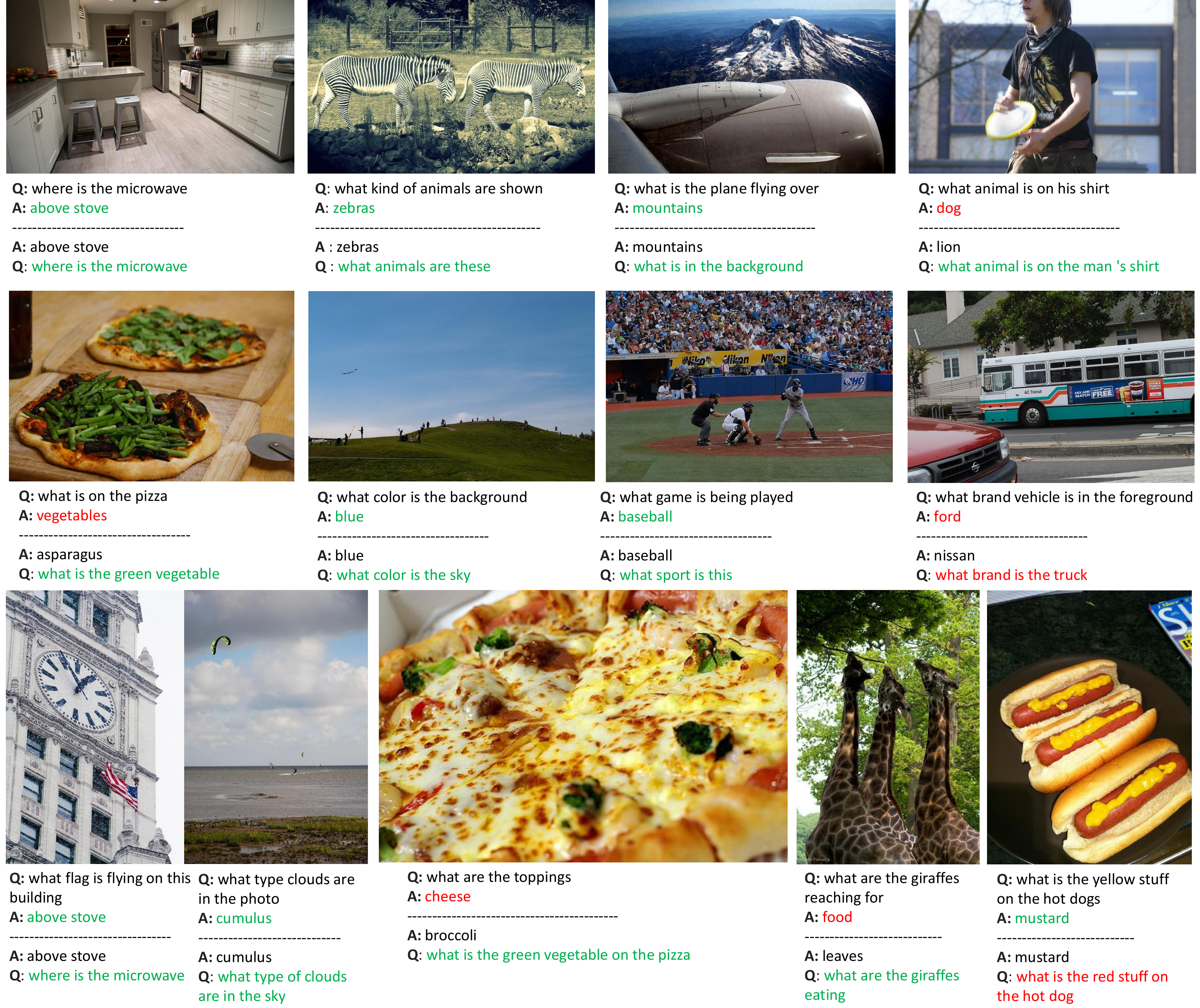}
	\end{center}
	\caption{Qualitative results of dual-trained VQA~(top) and VQG~(bottom) models. Labeled questions/answers are given to infer the counterpart. \textcolor{red}{Red} denotes the failure cases. \textcolor{green}{Green} indicates the correct ones. The question is viewed as \emph{correct} if it corresponds to the given answer. }
	\label{fig:qualitative}
\end{figure*}

\subsection{Analysis}

From the qualitative results, we can see that, our VQA and VQG models have the ability to recognize the image, understand the questions, and even know some uncommon words~(like \emph{cumulus}). The VQA model can associate the question and image to find the answer, while VQG model can generate  questions corresponding to the given answers although they are not identical to the labeled ones. 
Additionally, some typical failure cases are shown, including generating the wrong questions~(that are not corresponding the given answers), giving the wrong answer due to poor visual recognition ability or providing another reasonable answer but different from the label. Sometimes, questions are too hard to answer, e.g. \emph{to recognize the brand of vehicles}, where the question involves fine-grained categories and tiny objects.  But overall, the generated questions are similar to the human-annotated ones and correspond to the answers, which explains why they could serve as augmented data to improve VQA performance. 


\end{document}